\documentclass[sigplan,screen]{acmart}

\usepackage{multirow}
\usepackage{array}
\usepackage{graphicx} 
\usepackage{booktabs} 
\usepackage{subcaption}
\usepackage{enumitem}
\usepackage{threeparttable}

\AtBeginDocument{%
  \providecommand\BibTeX{{%
    \normalfont B\kern-0.5em{\scshape i\kern-0.25em b}\kern-0.8em\TeX}}}

\setcopyright{acmlicensed}
\copyrightyear{2018}
\acmYear{2018}
\acmDOI{XXXXXXX.XXXXXXX}

\acmConference[Conference acronym 'XX]{Make sure to enter the correct
  conference title from your rights confirmation emai}{June 03--05,
  2018}{Woodstock, NY}
\acmISBN{978-1-4503-XXXX-X/18/06}

\begin{document}

\title{Navigating Spatio-Temporal Heterogeneity:\\
A Graph Transformer Approach for Traffic Forecasting}



\author{Jianxiang Zhou, Erdong Liu, Wei Chen, Siru Zhong, Yuxuan Liang}
\authornote{corresponding author.}
\affiliation{%
  \institution{Hongkong University of Science and Technology (Guangzhou)}
  \country{}
  }
\email{jasonzhou314@gmail.com,eliu209@connect.hkust-gz.edu.cn, onedeanxxx@gmail.com,zhongsiru204@gmail.com, yuxliang@outlook.com}

\begin{abstract}
Traffic forecasting has emerged as a crucial research area in the development of smart cities. 
Although various neural networks with intricate architectures have been developed to address this problem, they still face two key challenges:
\textit{i)} Recent advancements in network designs for modeling spatio-temporal correlations are starting to see diminishing returns in performance enhancements.
\textit{ii)} Additionally, most models do not account for the spatio-temporal heterogeneity inherent in traffic data, \textit{i.e.,} traffic distribution varies significantly across different regions and traffic flow patterns fluctuate across various time slots.
To tackle these challenges, we introduce the Spatio-Temporal Graph Transformer (STGormer), which effectively integrates attribute and structure information inherent in traffic data for learning spatio-temporal correlations, and a mixture-of-experts module for capturing heterogeneity along spaital and temporal axes.
Specifically, we design two straightforward yet effective spatial encoding methods based on the graph structure and integrate time position encoding into the vanilla transformer to capture spatio-temporal traffic patterns. 
Additionally, a mixture-of-experts enhanced feedforward neural network (FNN) module adaptively assigns suitable expert layers to distinct patterns via a spatio-temporal gating network, further improving overall prediction accuracy. 
Experiments on real-world traffic datasets demonstrate that STGormer achieves state-of-the-art performance. 
The code is available at https://github.com/jasonz5/STGormer.
\end{abstract}



\maketitle

\section{Introduction}
\begin{figure*}[!htbp]
    \centering
    \begin{subfigure}{0.33\textwidth}
        \centering
        \includegraphics[width=\linewidth]{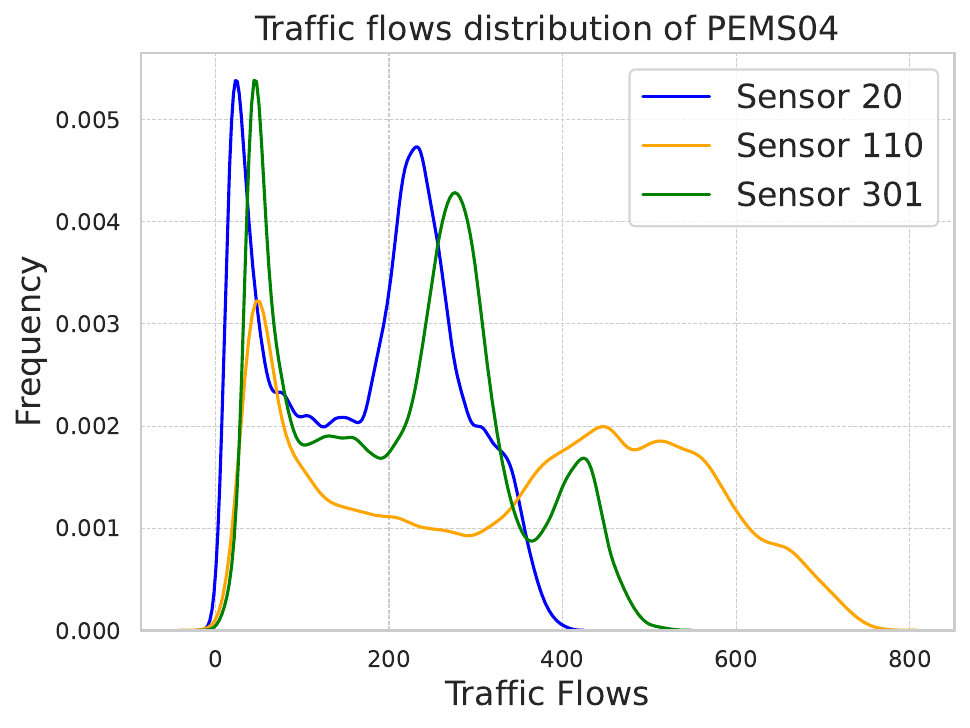}
        \caption{Traffic Distribution}
        \label{fig:1-motivation-d}
    \end{subfigure}
    \hfill
    \begin{subfigure}{0.33\textwidth}
        \centering
        \includegraphics[width=\linewidth]{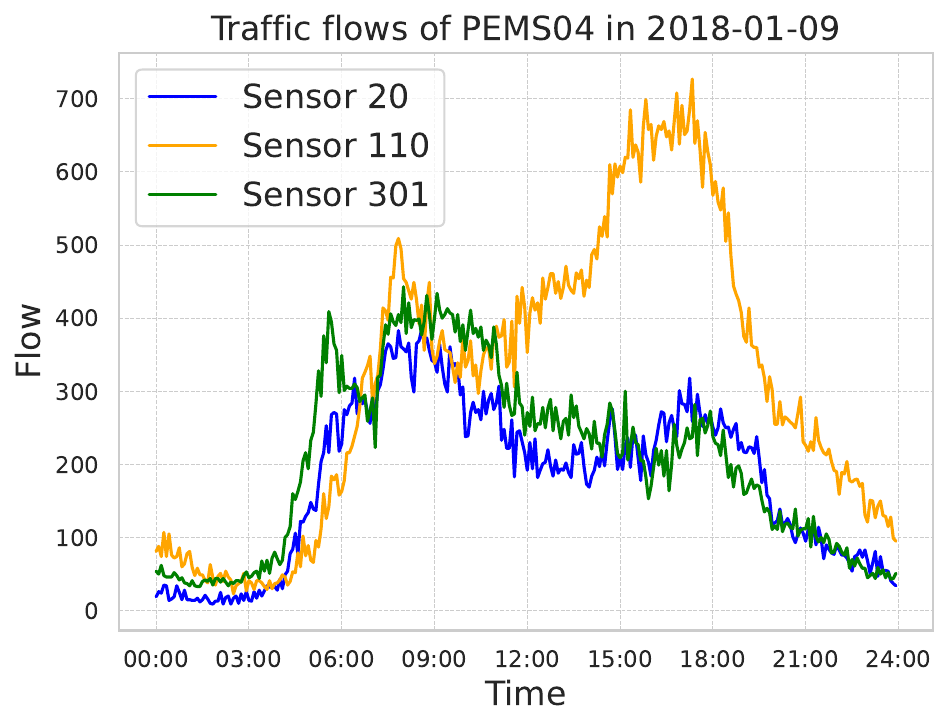}
        \caption{Spatial Heterogeneity}
        \label{fig:1-motivation-s}
    \end{subfigure}
    \hfill
    \begin{subfigure}{0.33\textwidth}
        \centering
        \includegraphics[width=\linewidth]{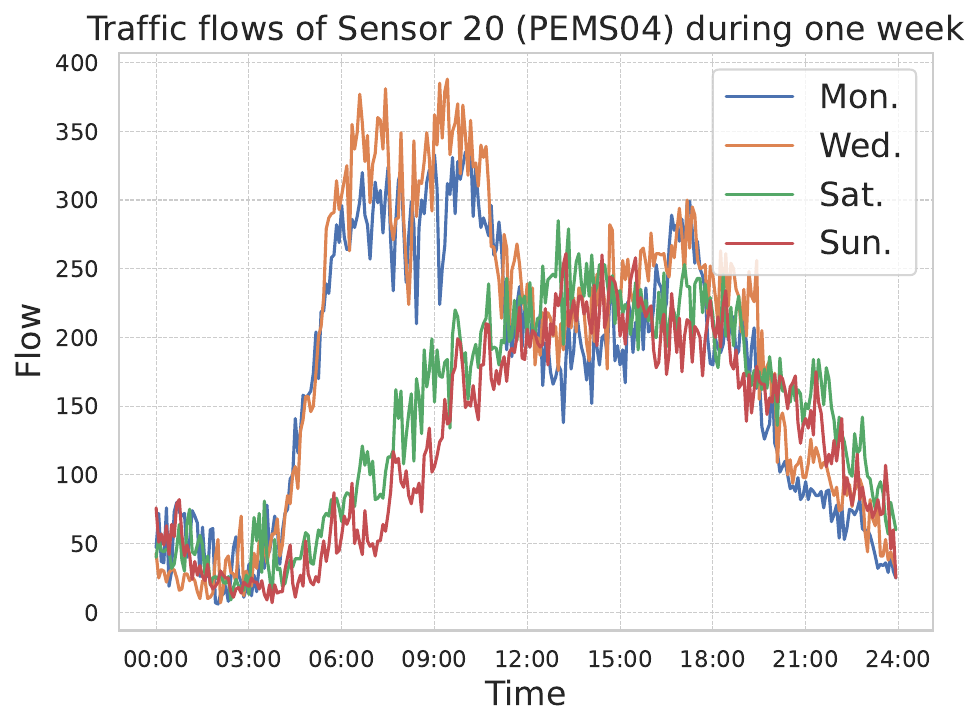}
        \caption{Temporal Heterogeneity}
        \label{fig:1-motivation-t}
    \end{subfigure}
    \caption{Illustration of spatio-temporal heterogeneity of traffic flow prediction.}
    \label{fig:1-motivation}
    \vspace{-1em}
\end{figure*}

Traffic forecasting~\cite{tedjopurnomo2020survey} has garnered substantial attention from both academic and industrial sectors due to its pivotal role in urban traffic management and planning. The goal is to predict future traffic patterns on road networks using historical data. 
This research area is marked by the need to decipher complex temporal and spatial patterns~\cite{guo2021hierarchical}, with the temporal aspect often exhibiting cyclical trends on a daily or weekly basis, and the spatial dimension characterized by the connectivity and interactions among various urban locations~\cite{shao2022pre}.
In recent years, various deep learning approaches have been developed to effectively capture the inherent spatio-temporal dependencies within traffic data~\cite{jin2023spatio}. 
Despite these advancements, current models still encounter two primary limitations.

The first shortcoming is the \textit{limited performance increment of continuously evolving complex architectures.}
To address the problems related to traffic forecasting, various deep learning models have been proposed in recent years. 
Among them, Spatial-Temporal Graph Neural Networks (STGNNs) have garnered significant attention for their accuracy in predictions~\cite{bai2019stg2seq}. They integrate Graph Neural Networks (GNNs)~\cite{kipf2016semi} with sequential models~\cite{sutskever2014sequence,yu2019review}, where the former captures dependencies among time series, the latter identifies temporal patterns. 
Other studies have incorporated attention models to enhance dynamic spatial modeling, taking advantage of no spatial constraints, as exemplified by GMAN~\cite{zheng2020gman} and STAEformer~\cite{liu2023spatio}. Although attention-based models fuse the spatio-temporal representation with one step, recent research has shown that they frequently struggle to generate meaningful attention maps, instead diffusing attention weights across all nodes~\cite{jin2022visual}. And these models often overlook spatial structural information embedded within graphs during attention computation.
Other sophisticated models are also explored for traffic forecasting, such as innovative graph convolutions across regional grids~\cite{guo2019attention}, algorithms for learning graph structures~\cite{wu2019graph}, and efficient attention mechanisms~\cite{zhou2021informer}.
However, these enhancements in network architectures have begun to yield diminishing returns in terms of performance improvements. This observation has catalyzed a shift in focus from the complexity of model designs to the development of effective representation techniques for the underlying data~\cite{gao2023spatio}.

The second limitation is the \textit{lack of modeling spaital and temporal heterogeneity}. This refers to the significant variations in traffic distribution across different regions and the fluctuations in traffic flow patterns at various times.
For example, as shown in Figure~\ref{fig:1-motivation-s}, sensors $20$ and $301$ exhibit very similar patterns over the same time frame due to their proximity within the traffic network, while sensor $110$ displays distinct trends. We can also observe their similar relations of traffic flow distributions from Figure~\ref{fig:1-motivation-d}.
Moreover, Figure~\ref{fig:1-motivation-t} reveals distinct weekday and weekend traffic patterns particularly during peak hours. Weekdays show a pronounced morning rush hour peak, while weekends exhibit a more uniform distribution throughout the day without significant peaks, highlighting the temporal heterogeneity in weekly patterns.
Current methodologies predominantly utilize a unified parameter space for analyzing all traffic data, which often fails to accurately capture the spatio-temporal heterogeneity in the latent embedding space. 
In addition, recent trends have seen the integration of self-supervised learning techniques to dynamically model various traffic patterns~\cite{ji2023spatio,shao2022pre,zhang2023spatial}. However, these approaches typically involve complex model architectures and intricate training paradigms, which may complicate their practical deployment.

To effectively model spatio-temporal correlations and heterogeneity, we propose \underline{S}patio-\underline{T}emporal \underline{G}raph Transf\underline{ormer} (STGormer) for traffic forecasting, and achieves state-of-the-art performance on three real-world traffic datasets. 
To model spatial and temporal correlations without relying on complex architectures, our STGormer is built over a Transformer model which incorporates several attribute and structure embeddings. Specifically, we add temporal positional information and degree centrality into the node features of the input layer. Additionally, we integrate structural information of the graph into the spatial attention layer, enabling the model to effectively capture spatial and temporal dependencies in traffic data.
To address spatial and temporal heterogeneity, our approach includes a mixture-of-experts enhanced FNN module. This module adaptively assigns appropriate expert layers to distinct patterns through a spatio-temporal gating network, which also reveals complex spatio-temporal relationships within the data.

Our major contributions can be summarized as follows:
\begin{itemize}[leftmargin=*]
    \item We propose a novel spatio-temporal graph transformer model called STGormer for traffic forecasting. Specifically, the model incorporates attribute and structure information for effectively modeling spatio-temporal correlations, and a mixture-of-experts module for capturing spatial and temporal heterogeneity in traffic flow prediction. 
    \item By integrating simple yet effective structural encoding from spatio-temporal graphs, the model significantly enhances forecasting accuracy and improves the transformer's ability to comprehend graph structure information.
    \item Extensive empirical studies conducted on three real-world datasets illustrate that STGormer substantially surpasses baseline models in traffic forecasting accuracy, while also demonstrating superior ability to modeling traffic patterns.
\end{itemize}

\section{Preliminary}
\subsection{Formulation}

\textbf{Definition 1. \textit{Spatio-temporal Graph:}}
We define traffic flow data as a spatio-temporal graph $G = (V, E, A, X)$, where $ V $ is a set of all roads or sensors in road networks with $ |V| = N $, $ E $ is a set of edges representing the connectivity between nodes, and $ A \in \mathbb{R}^{N \times N} $ is a matrix representing the topology of graph. And $ X \in \mathbb{R}^{T \times N \times C} $ defines traffic flows, where $T$ is the number of time steps, $N$ is the number of variables, and $C$ is the number of input features.

\noindent\textbf{Definition 2. \textit{Traffic Forecasting:}}
We formulate our forecasting problem as predicting future $ T' $ traffic flows based on $ T $ historical input data:
\begin{equation}
\left[ X^{(t-T+1)}, \ldots, X^{(t)} \right] \xrightarrow{f(\cdot)} \left[ X^{(t+1)}, \ldots, X^{(t+T')} \right]
\end{equation}
where $ X^{(i)} \in \mathbb{R}^{N \times C} $. We aim to train the mapping function 
$ f(\cdot) : \mathbb{R}^{T \times N \times C} \rightarrow \mathbb{R}^{T' \times N \times C} $
which predicts the next $ T' $ steps based on the given $ T $ observations.

\noindent\textbf{Definition 3. \textit{Expert:}}
An expert is a specialized sub-network within a base model that is designed to analyze spatio-temporal dependencies. This sub-network is combined with a gated network, which allows for targeted processing of distinct spatio-temporal patterns.

\begin{figure*}[!t]
  \centering
  \includegraphics[width=\textwidth]{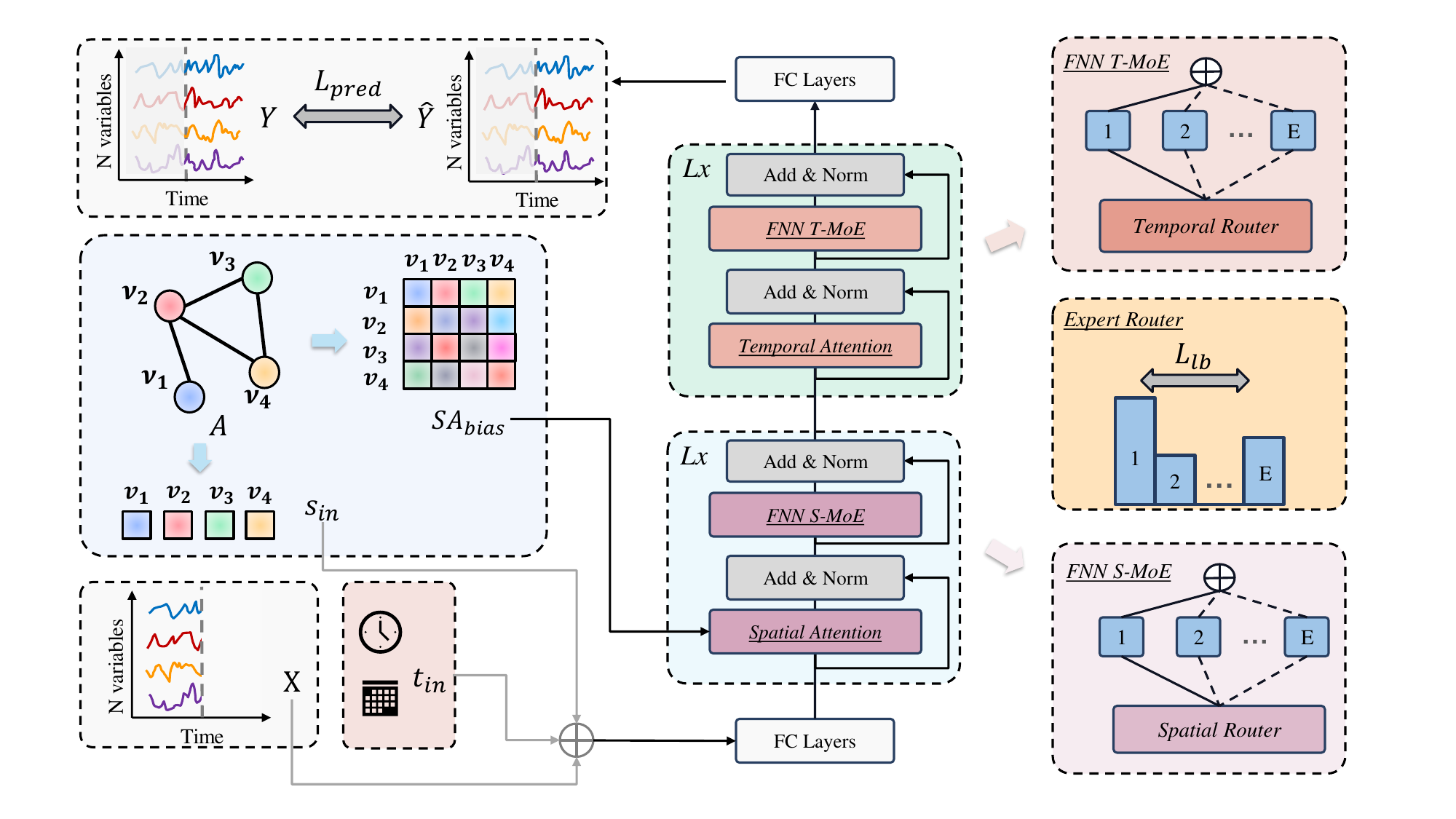}
  \caption{The framework of Spatio-Temporal Graph Transformer (STGormer).}
  \label{fig:4-framework}
\end{figure*}

\subsection{Related Works}

\subsubsection{Spatio-Temporal Forecasting}
Spatio-temporal forecasting has been studied for decades, aiming to predict future states by analyzing historical data.
Initially, conventional time series models dominated the field, but their limited capacity to handle intricate temporal and spatial correlations led to the adoption of more advanced methods. The introduction of Convolutional Neural Networks (CNNs)~\cite{oord2016wavenet} and Recurrent Neural Networks (RNNs)~\cite{lv2018lc}  marked a significant improvement, offering better handling of the complexities in traffic time series data.
However, the real breakthrough came with Spatial-Temporal Graph Neural Networks (STGNNs), which combined the strengths of Graph Neural Networks (GNNs)~\cite{kipf2016semi} with sequential models~\cite{sutskever2014sequence,yu2019review}, enabling the joint modeling of spatial and temporal dependencies~\cite{wu2019graph, xu2020spatial,deng2023learning}. 
Several STGNNs models have been proposed in recent years, achieving remarkable results in urban traffic flow prediction tasks, such as Graph WaveNet~\cite{wu2019graph}, STGCN~\cite{yu2017spatio}, DCRNN~\cite{li2017diffusion} and AGCRN~\cite{bai2020adaptive}. 

Moreover, attention mechanism~\cite{vaswani2017attention} has also become increasingly popular due to its effectiveness in modeling the dynamic dependencies in traffic data, such as GMAN~\cite{zheng2020gman} and ASTGCN~\cite{guo2019attention} to model spatio-temporal relations.
Attention-based models have shown impressive performance but struggle to adequately capture the inherent spatial structure of traffic flow data. This study introduces several straightforward yet effective structural encoding techniques to enhance the vanilla transformer, allowing it to effectively capture spatio-temporal correlations.

\subsubsection{Mixture of Experts}
The Mixture-of-Experts (MoEs) technique has gained significant attention for enabling efficient and accurate handling of complex data since its introduction to language modeling~\cite{shazeer2017outrageously}. MoEs have been applied across diverse machine learning fields, including computer vision~\cite{dryden2022spatial}, natural language processing~\cite{du2022glam}, and vision-language modeling~\cite{lin2024moe}. 
In recent years, MoEs have also been utilized in traffic forecasting. For instance, TESTAM~\cite{lee2024testam} employs various spatial modeling methods as experts for more flexible spatial modeling, while ST-ExpertNet~\cite{wang2022st} uses three spatial experts to handle flow patterns related to work, entertainment, and commuting.  
STGormer distinguishes itself by integrating MoE-enhanced FNN at both spatial and temporal levels, emphasizing expert specialization to address spatio-temporal heterogeneity effectively.

\section{Methodology}
The Spatio-Temporal Graph Transformer (STGormer) framework comprises a spatio-temporal encoding layer, MoE enhanced transformers along the spatial and temporal axes, and a regression layer for traffic flow prediction. The architecture of STGormer is illustrated in Figure~\ref{fig:4-framework}.

\subsection{Spatio-Temporal Encoding Layer}
To capture spatio-temporal representations, we introduce a Spatio-temporal Encoding Layer that integrates both spatial and temporal static context within the traffic flow graph. 
For embedding temporal position information, we incorporate features such as sequential time positions or timestamp information. 
For spatial input encoding, we utilize the degree centrality which represents the importance of each node in traffic network. 
These spatio-temporal attribute and structure embeddings are incorporated into the traffic flow data before being input into the MoE-enhanced spatio-temporal transformers.

\subsubsection{Temporal Input Encoding}
Temporal features, such as the time of day and the day of the week, serve as global position indicators with inherent periodicity. Instead of relying on normalized temporal features, we leverage Time2Vec embeddings~\cite{mehran2019time2vec} to effectively capture both periodic and linear aspects. Specifically, for a given temporal static context $ t $, we compute the temporal input encoding $ t_{in} \in \mathbb{R}^{T\times d} $ using a $ d $-dimensional embedding vector $ v(t) $. This vector is parameterized by learnable parameters $ w_i $ and $ b_i $ for each dimension $ i $, as illustrated below:
\begin{equation}
    t_{in}[i] = 
        \begin{cases} 
        w_i * v(t)[i] + b_i, & \text{if } i = 0 \\
        F(w_i * v(t)[i] + b_i), & else
        \end{cases}
\end{equation}
where $F$ denotes a periodic activation function. By incorporating Time2Vec embeddings, the model effectively harnesses the temporal information associated with labels.

\subsubsection{Spatial Input Encoding}
In the layer dedicated to capturing spatio-temporal representations, we employ an attention-based model. This approach facilitates the simultaneous modeling of sequential traffic patterns over various time steps and the geographic relationships between different spatial regions. 
While attention-based models are renowned for their effectiveness in time-series forecasting, they encounter difficulties in modeling spatial representations. This is due to their model structure, which lacks inherent incorporation of spatial prior knowledge from graphs. To overcome these challenges, we propose two straightforward yet effective methods for embedding spatial structure information into the input encoding layer and the spatial attention layer.

In the spatio-temporal encoding layer, we leverage the structural properties of graphs by incorporating degree centrality as the spatial input encoding $s_{in}$. Degree centrality, defined as the number of connected neighboring nodes within a predetermined adjacency matrix, quantifies the significance of a node within the graph. 
For example, in a transportation network, nodes representing transportation hubs with numerous connecting routes are pivotal and therefore deemed more critical due to their influence on flow dynamics. 
Despite its importance, this aspect of node significance is typically neglected in the computation of attention within Transformer models. We propose to integrate this information as a valuable signal to enhance model performance. Specifically, for a traffic flow graph, the spatial input encoding $s_{in}$ is computed as follows:
\begin{equation}
    s^{(in)} = z^-_{\deg^-(v_i)} + z^+_{\deg^+(v_i)},
\end{equation}
where $ z^-, z^+ \in \mathbb{R}^d $ represent learnable embedding vectors that correspond to the indegree $ \deg^-(v_i) $ and outdegree $ \deg^+(v_i) $ of a node $ v_i $, respectively. In the case of undirected graphs, both $ \deg^-(v_i) $ and $ \deg^+(v_i) $ converge to $ \deg(v_i) $. By encoding node centrality into the input, our model enables the softmax attention mechanism to recognize and leverage the importance of nodes in both the query and key vectors, thereby capturing not only the semantic correlations but also the centrality signals within the graph.

\subsubsection{Fused Spatio-Temporal Input Encoding}
Prior to processing through the spatio-temporal Transformer, the traffic flow embeddings $X$ are merged with the temporal input encoding $t_{in}$ and spatial input encoding $s_{in}$. This concatenated vector is then mapped onto a hidden dimension $d$ as follows:
\begin{equation}
    H = FC( X \; \| \; t_{in} \; \| \; s_{in}),
\end{equation}
where $H \in \mathbb{R}^{T \times N \times D}$ represents the hidden spatio-temporal representation. This step is critical as it integrates both spatial and temporal static context, enhancing the Transformer’s capability to analyze complex dependencies in traffic flow data effectively.

\subsection{MoE-enhanced Spatio-Temporal Transformer}
For modeling spatio-temporal representations, we employ the Attention mechanism, focusing exclusively on either the spatial or temporal dimensions to capture spatio-temporal correlations. Specifically for spatial attention, we utilize a spatial attention bias matrix derived from the shortest path of nodes in the graph topology. The outputs of attention layer are then fed into a Feedforward Neural Network (FNN)-based Mixture-of-Experts (MoE) model, comprising $E$ duplicated FNNs, each acting as an "expert" for learning spatio-temporal heterogeneity.
The final spatio-temporal features of MoE-enhanced transformers are passed through a regression layer for the ultimate task of prediction, effectively leveraging the intricate spatio-temporal dynamics captured by the network.

\subsubsection{Spatio-temporal Attention}
We apply attention layers along the temporal and spatial axes to capture intricate traffic relations. The computation of attention aligns with that used in Transformers, employing Multi-head Self Attention (MSA) in the temporal or spatial dimension. However, there are slight modifications for spatial attention, integrating spatial structural information into the attention mechanism.

Given the hidden spatio-temporal representation $ H \in \mathbb{R}^{T \times N \times D} $, where $ T $ denotes the time frames and $ N $ represents spatial nodes, serving as the input for the self-attention module. The input $ H $ is projected by three matrices $ W_Q , W_K , W_V \in \mathbb{R}^{d \times d} $ to derive the corresponding query, key, and value matrices as follows:
\begin{equation}
    Q = HW_Q, \quad K = HW_K, \quad V = HW_V,
\end{equation}

Then we proceed to compute the self-attention score $A $ and derive the hidden immediate spatio-temporal representation $H' $ as follows:
\begin{equation}
    A = \frac{QK^T}{\sqrt{d}}, \quad H' = \textit{Softmax}(A)V,
\end{equation}
Concerning the self-attention score $A $, for temporal attention, $A_t \in \mathbb{R}^{N \times T \times T} $ is computed as described above. However, for spatial attention, the calculation is slightly different, incorporating spatial structural information.

To encode the spatial information of a graph within the attention mechanism, we employ the spatial attention bias matrix inspired by Graphormer~\cite{ying2021transformers}. Specifically, for any graph $G $, we consider the shortest path distance (SPD) between two connected nodes. If not connected, we set the SPD output to a special value, $-1 $ in this paper. We calculate the spatial attention bias matrix $SA_{bias} $ by element-wise embedding the SPD matrix with a learnable scalar. Denoting $A_s \in \mathbb{R}^{T \times N \times N} $ as the spatial self-attention score matrix, we have:
\begin{equation}
    A_{s} = \frac{QK^T}{\sqrt{d}} + SA_{bias},
\end{equation}
\begin{equation}
    SA_{bias} = \phi( \textit{SPD}(G) ),
\end{equation}
where $SPD $ represents the algorithm for calculating the shortest path distance between nodes, $SA_{bias} \in \mathbb{R}^{N\times N} $, $G $ is the adjacency matrix of the traffic flow graph, and $\phi $ is an element-wise learnable scalar shared across all spatial attention layers.

After obtaining the spatio-temporal intermediate hidden representations $H'$, we pass them to the spatio-temporal router network, which will be elaborated in the subsequent part.

\subsubsection{Spatio-Temporal Router}

To bolster the model's ability to capture spatio-temporal heterogeneity, we leverage a Mixture-of-Experts (MoEs) enhanced FNN mudule.

\textbf{Gating Network}. Employing a fine-grained routing strategy for expert selection, we utilize the representations derived from the spatial and temporal attention layers as inputs to the gating network. This network computes the weight distribution of experts using a straightforward MLP with a softmax function:
\begin{equation}
    G( H') = \textit{Softmax}(\text{MLP}(H')),
\end{equation}

\textbf{Output Aggregation}. Then we pass the representations derived from the spatial and temporal attention layers as inputs to experts layer and integrates all the predictions from each expert. The calculation is formulated as below:
\begin{equation}
    H = \sum_{i=1}^{E} G_i(H') \odot E_i(H'),
\end{equation}
Where $G(H')$ and $E_i(H')$ denote the output of the gating network and the output of the $i$-th expert network for a given input $H'$. 
In our setup, we employ two distinct router networks: $G_s$ and $G_t$, dedicated to handling spatial and temporal relations respectively. In this architecture, each expert is represented by an individual neural network, each operating with its own set of parameters to learn unique spatio-temporal patterns.

\textbf{Load Balancing}. 
In our model's training process, we encountered a phenomenon previously observed in other related works ~\cite{shazeer2017outrageously} where the gating network tends to assign disproportionately higher weights to a few experts. To mitigate this issue, we incorporated the concept of load balancing. This strategy aims to ensure a more equitable distribution of the utilization load among the experts. Mathematically, we define the load balancing loss term as:
\begin{equation}
    L_{lb} = \frac{1}{E}\sum_{i=1}^{E} f_i^2,
\end{equation}
where $E$ represents the number of experts, and $f_i$ denotes the fraction of probability mass allocated to expert $i$.

\subsection{Model Training}

For training the STGormer model, we utilize the final representation $H_{\text{final}} \in \mathbb{R}^{T \times N \times D}$ generated by the MoE-enhanced Spatio-Temporal Transformer. This representation is fed into a regression layer to enable the prediction of traffic flow for future time steps, expressed as:
\begin{equation}
    \hat{Y} = \textit{FC}(H_{\text{final}}),
\end{equation}

To optimize the model, we minimize the mean absolute error (MAE) between $\hat{Y}$ and the ground truth $Y$. The total loss function is defined as:
\begin{equation}
    L(\theta) = \text{MAE}_{\text{loss}}(Y - \hat{Y}) + \alpha L_{\text{lb}},
\end{equation}
where $\theta$ represents all the learnable parameters in STGormer, and $\alpha$ is a parameter used to balance the influence of the load balancing loss term.

\section{Experiment}
\renewcommand{\arraystretch}{1.5}
\begin{table*}[!h]
    \centering
    \caption{Statistics of Datasets.}
    \begin{tabular}{c|c|c|c|c|c}
        \hline\hline 
\textbf{Dataset} & \textbf{Domain} & \textbf{Temporal Duration} & \textbf{Temporal Interval} & \textbf{Nodes}&\textbf{Edges}\\ \hline\hline 
NYCBike1 & Bike usage & 20140401-20140930 & One hour & 128 &442\\ 
NYCBike2 & Bike usage & 20160701-20160829 & Half an hour & 200 &712\\ 
NYCTaxi & Taxi OD & 20160101-20160229 & Half an hour & 200 &712\\ 
\hline \hline 
    \end{tabular}
    \label{tab:datasets}
\end{table*}

\subsection{Experimental Settings}

\subsubsection{Datasets and Metrics}
To thoroughly evaluate the proposed STGormer model, we conduct extensive experiments on three real-world traffic datasets, as listed in Table~\ref{tab:datasets}: NYCTaxi, NYCBike1~\cite{yao2019revisiting}, NYCBike2~\cite{lin2019deepstn+}.
For these datasets, we predict the flows for the next time step using the previous 2-hour flows and the previous 3-day flows around the predicted time, ensuring fair comparison with previous work~\cite{ji2023spatio}. 
All datasets are divided into training, validation, and test sets with a ratio of 7:1:2.

We utilize three commonly used evaluation metrics in the field of traffic prediction to provide a fair assessment of the prediction performance: Mean Absolute Error (MAE), Root Mean Squared Error (RMSE), and Mean Absolute Percentage Error (MAPE). MAE and RMSE measure absolute prediction errors, while MAPE measures relative prediction errors. For all these metrics, smaller values indicate better performance.

\subsubsection{Implementation Details.}
We implement the model using the PyTorch toolkit on a Linux server equipped with an NVIDIA GeForce RTX 3090 GPU. 
The embedding dimension $D$ is set to 64. 
The number of both spatial and temporal MoE-enhanced transformer blocks is set to 3. 
The number of attention heads is set to 4. Each spatial and temporal router network contains 6 experts. 
Optimization is performed with Adam optimizer with the learning rate decaying from 0.001, and the batch size is set to 32. We apply an early-stop mechanism if the validation error converges within 25 continuous steps.
Hyperparameters were chosen based on commonly used values and adjusted through preliminary experiments, without extensive ablation or hyperparameter optimization.

\subsubsection{Baselines.}
In this study, we compare our proposed method against several widely used baselines in the field. 
Historical Average (HA) and Support Vector Regression (SVR)~\cite{smola2004tutorial} are traditional statistical methods. For deep learning methods, we include ST-ResNet~\cite{zhang2017deep}, STGCN~\cite{yu2017spatio}, GMAN~\cite{zheng2020gman}, Graph WaveNet~\cite{wu2019graph}, and AGCRN~\cite{bai2020adaptive}, each demonstrating significant capabilities in capturing spatial and temporal dependencies within datasets. Additionally, we select STID~\cite{shao2022spatial}, 
STAEformer~\cite{liu2023spatio}, and ST-SSL~\cite{ji2023spatio} as recent state-of-the-art developments to explore advanced neural network architectures and input embedding methods.

\subsection{Main Results}
\renewcommand{\arraystretch}{1.5}
\setlength{\tabcolsep}{2pt}
\begin{table*}[!ht]
\footnotesize
\small
\centering
\caption{Model comparison on traffic datasets in terms of MAE, RMSE and MAPE (\%).}
\begin{threeparttable}
\begin{tabular}{cc|c|c c|c c c c c|c c c|c}
\hline

\multicolumn{2}{c|}{Datasets}                                                                                                  & Metric & HA    & SVR   & ST-ResNet & STGCN & GMAN  & GWNet & AGCRN & STID  & STAEformer & ST-SSL & \textbf{STGormer} \\ \hline
\multicolumn{1}{c}{\multirow{3}{*}{NYCBike1}} & \multirow{3}{*}{\begin{tabular}[c]{@{}c@{}}Horizon 1\\ (60 min)\end{tabular}}  & MAE    & 7.89  & 7.63  & 5.64      & 5.46  & 6.97  & 5.45  & 5.32  & 6.66  & 5.14       & 5.17   & \textbf{4.99}              \\ 
\multicolumn{1}{c}{}                          &                                                                                & RMSE   & 12.13 & 11.35 & 8.35      & 8.03  & 10.11 & 8.04  & 7.86  & 10.33 & 7.51       & 7.75   & \textbf{7.36}              \\ 
\multicolumn{1}{c}{}                          &                                                                                & MAPE   & 28.42 & 26.41 & 25.91     & 27.31 & 33.23 & 23.60 & 26.11 & 28.26 & \textbf{22.86}      & 23.92  & 23.24             \\ \hline
\multicolumn{1}{c}{\multirow{3}{*}{NYCBike2}} & \multirow{3}{*}{\begin{tabular}[c]{@{}c@{}}Horizon 1\\ (30 min)\end{tabular}}  & MAE    & 12.76 & 12.15 & 5.85      & 5.07  & 5.11  & 5.13  & 4.99  & 5.39  & 4.84       & 4.86   & \textbf{4.79}              \\ 
\multicolumn{1}{c}{}                          &                                                                                & RMSE   & 18.95 & 18.59 & 8.84      & 7.72  & 7.84  & 7.97  & 7.70  & 8.58  & \textbf{7.24}       & 7.41   & 7.48              \\ 
\multicolumn{1}{c}{}                          &                                                                                & MAPE   & 45.62 & 44.22 & 31.33     & 26.28 & 27.07 & 23.50 & 26.66 & 23.89 & 21.72      & 22.04  & \textbf{21.18}             \\ \hline
\multicolumn{1}{c}{\multirow{3}{*}{NYCTaxi}}  & \multirow{3}{*}{\begin{tabular}[c]{@{}c@{}}Horizon 1\\ (30 min)\end{tabular}}  & MAE    & 48.62 & 46.94 & 12.08     & 11.74 & 13.58 & 10.81 & 11.00 & 15.56 & 10.72      & 10.73  & \textbf{10.42}             \\ 
\multicolumn{1}{c}{}                          &                                                                                & RMSE   & 83.55 & 82.98 & 21.76     & 21.01 & 24.85 & 19.45 & 20.06 & 29.97 & \textbf{19.13}      & 19.20  & 19.56             \\ 
\multicolumn{1}{c}{}                          &                                                                                & MAPE   & 65.35 & 64.58 & 24.62     & 20.89 & 22.35 & 18.34 & 18.60 & 21.10 & 15.94      & 16.12  & \textbf{15.83}             \\ \hline

\end{tabular} 
\end{threeparttable}
\label{tab:main_results}
\end{table*}

Table~\ref{tab:main_results} demonstrates that our STGormer framework exhibits enhanced performance across a majority of metrics on all datasets, affirming the efficacy of our approach. 
Please note that discrepancies in baseline results for the NYCBike1/2 and NYCTaxi datasets stem from our use of ST-SSL's evaluation methods, specifically calculating prediction errors for values that exceed a non-zero threshold in traffic flow.
Other spatio-temporal models also deliver strong results, attributing their success to adept handling of spatio-temporal dependencies. Conversely, traditional non-deep learning techniques such as HA and SVR are the least effective, struggling to model the complex non-linear spatial and temporal dynamics prevalent in real-world datasets.
In summary, the STGormer framework significantly advances the state-of-the-art in traffic forecasting, demonstrating its ability to capture the intricate spatio-temporal dynamics.

\subsection{Ablation Study}
\begin{figure}[!htbp]
  \centering
  \includegraphics[width=0.48\textwidth]{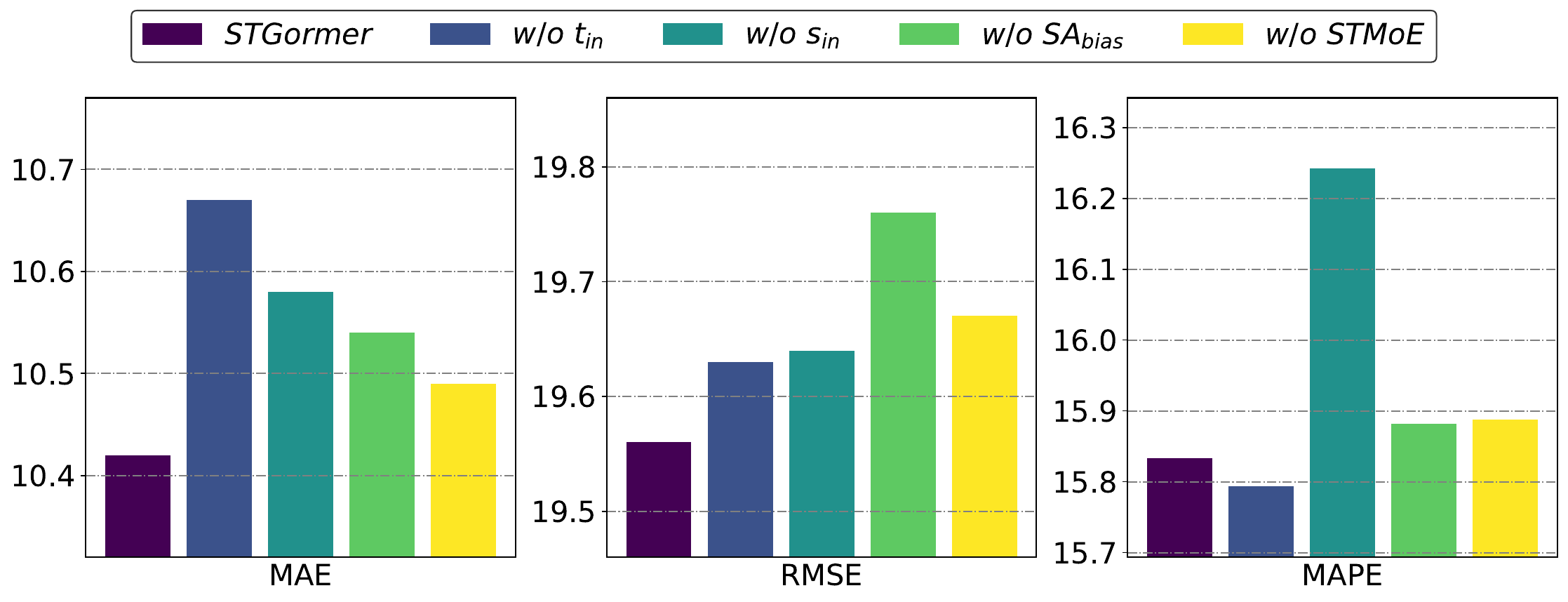}
  \caption{Ablation study on NYCTaxi.}
  \label{fig:5-ablation-study}
\end{figure}

To evaluate the effectiveness of STGormer, we conduct ablation studies with four variants of our model as follows:

\begin{itemize}
    \item \textbf{w/o $t_{in}$}: This variant omits the temporal input encoding $t_{in}$, which serves as temporal position indicators.
    \item \textbf{w/o $s_{in}$}: This variant omits the spatial input encoding $s_{in}$, which indicates the importance of each node in the graph.
    \item \textbf{w/o $SA_{bias}$}: This variant removes the spatial attention bias matrix $SA_{bias}$, which is designed to integrate spatial relationships into the attention layer.
    \item \textbf{w/o STMoE}: This variant disables the MoE-enhanced feedforward neural network, relying solely on the original point-wise feedforward network to capture spatial and temporal representations.
\end{itemize}

Figure~\ref{fig:5-ablation-study} illustrates the impact of various components on the performance of our model when tested on the NYCTaxi 
dataset. 
The temporal input encoding $t_{in}$ is essential for capturing both periodic and linear aspects of temporal features. 
Furthermore, the significant performance degradation observed upon removing the spatial input encoding $s_{in}$ and the spatial attention bias matrix $SA_{bias}$ demonstrates that our proposed spatial structure embeddings effectively model the inherent spatial patterns in traffic data and enhance the transformers' capability to model spatio-temporal graph data. 
The \textit{w/o STMoE} variant underscores the crucial role of the Mixture of Experts (MoE) in enhancing the feedforward network's ability to capture spatio-temporal patterns, where specialized routers direct inputs to the appropriate experts, thereby improving the management of complex traffic patterns. 
Overall, the results underscore the value of our proposed spatio-temporal graph transformer for traffic forecasting.

\subsection{Hyper-parameter Study}
\begin{figure}[!htbp]
    \centering
    \begin{subfigure}{0.48\textwidth}
        \centering
        \includegraphics[width=\linewidth]{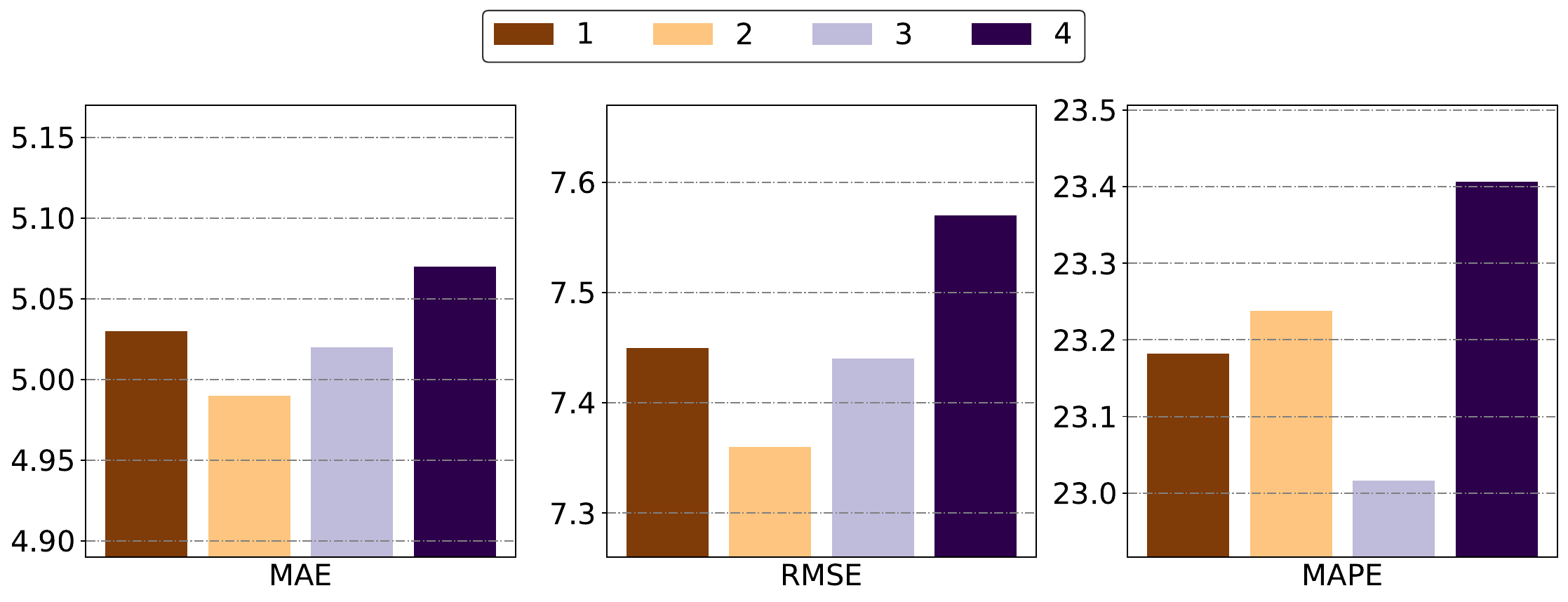}
        \caption{Impact of blocks number.}
        \label{fig:5-layers-num}
    \end{subfigure}
    \begin{subfigure}{0.48\textwidth}
        \centering
        \includegraphics[width=\linewidth]{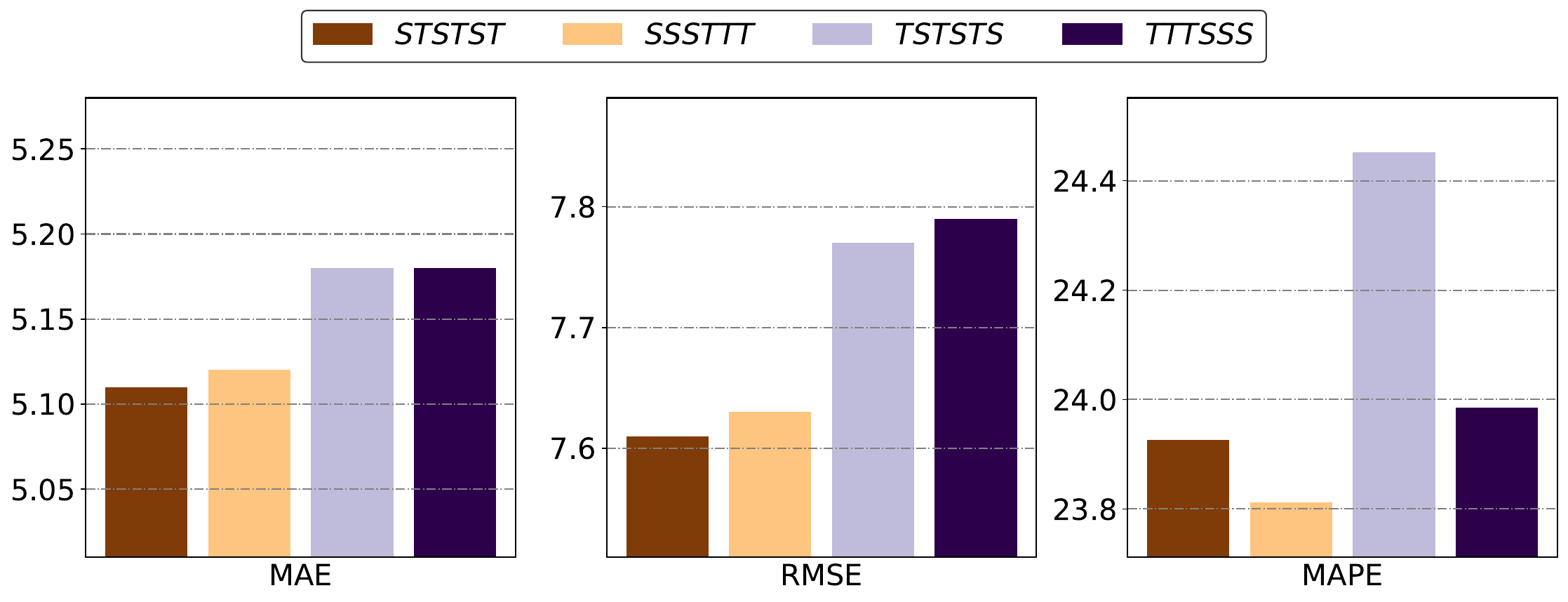}
        \caption{Impact of blocks order.}
        \label{fig:5-layers-order}
    \end{subfigure}
    \caption{Hyper-parameter study on NYCBike1.}
    \label{fig:5-hyper}
\end{figure}

We conduct experiments to analyze the impacts of two hyperparameters: the number of blocks in each spatial and temporal transformer, and the order of these blocks. We present our results on the NYCBike1 dataset.

\textbf{Blocks Number.} We first conduct hyperparameter study by varying the number of transformer blocks, setting the count equally for both spatial and temporal components to values in $\{1, 2, 3, 4\}$. Our analysis identifies an optimal balance, suggesting a delicate interplay between model complexity and performance efficacy, as shown in Figure~\ref{fig:5-layers-num}. Configurations with fewer blocks than the optimal value tend to simplify the model complexity and diminish its predictive capabilities. Conversely, a higher number of blocks may lead the model to overfit, capturing noise rather than meaningful patterns and thus degrading its generalization performance on unseen data.

\textbf{Blocks Order.} We extend our examination to the impact of varying the arrangement of spatial and temporal transformer blocks, focusing on configurations such as \textit{SSSTTT} and \textit{STSTST}, as well as their temporal-first counterparts, \textit{TTTSSS} and \textit{TSTSTS}. Our results, as detailed in Figure~\ref{fig:5-layers-order}, demonstrate that configurations prioritizing spatial modeling consistently yield the lowest values across all evaluation metrics compared to those beginning with temporal modeling. This finding underscores the importance of spatial processing as a foundational step, which significantly enhances the synergy between the spatial and temporal dimensions. By starting with spatial modeling, the model is better equipped to capture effective spatio-temporal representations, thereby improving its predictive accuracy in traffic forecasting tasks.

\subsection{Qualitative Study}
\begin{figure}[!htbp]
    \centering
    \begin{subfigure}[b]{0.48\linewidth}
        \centering
        \includegraphics[width=0.50\linewidth]{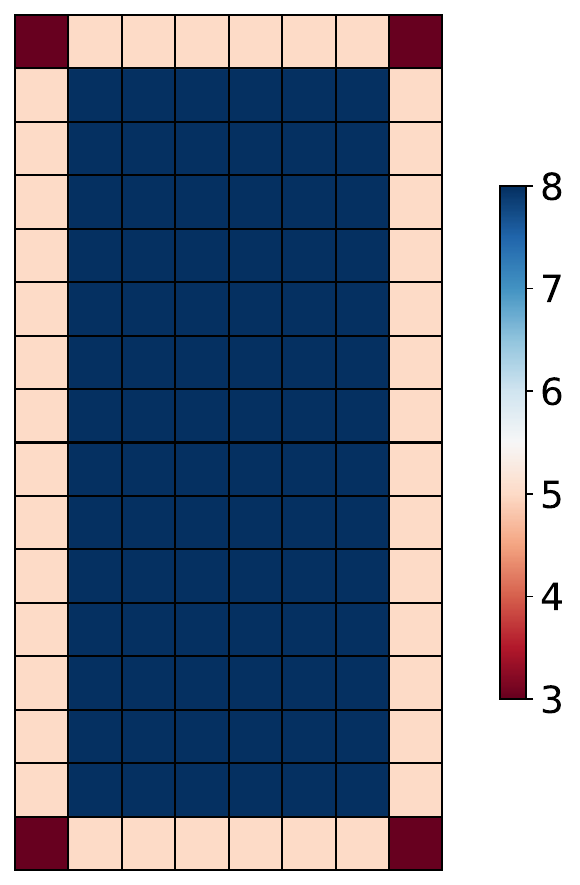}
        \caption{Visualization of $s_{in}$.}
        \label{fig:5-case-s_in}
    \end{subfigure}
    \begin{subfigure}[b]{0.48\linewidth}
        \centering
        \includegraphics[width=0.50\linewidth]{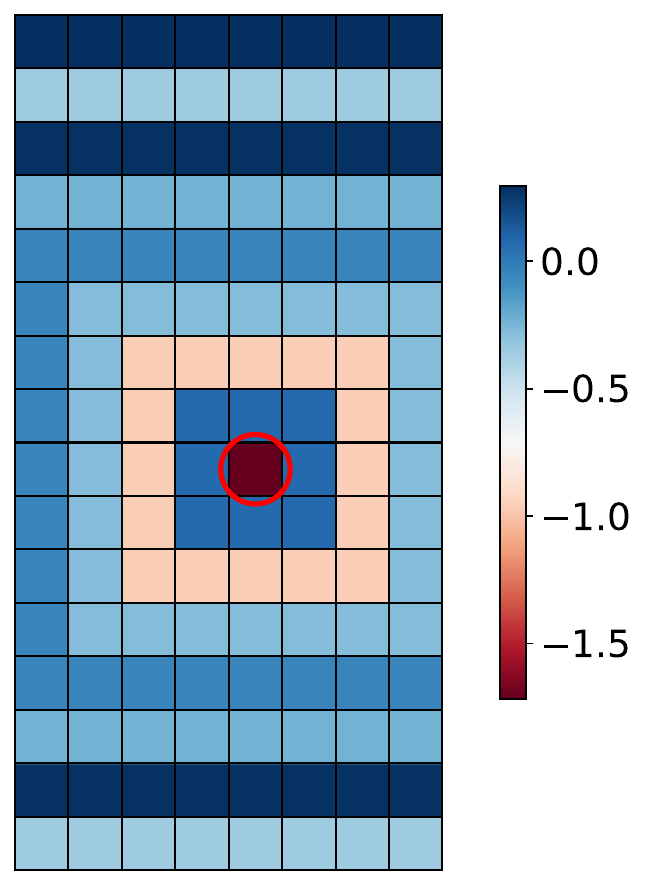}
        \caption{Visualization of $SA_{bias}$.}
        \label{fig:5-case-sa_bias}
    \end{subfigure}
    \begin{subfigure}[b]{0.48\textwidth}
        \centering
        \includegraphics[width=0.96\linewidth]{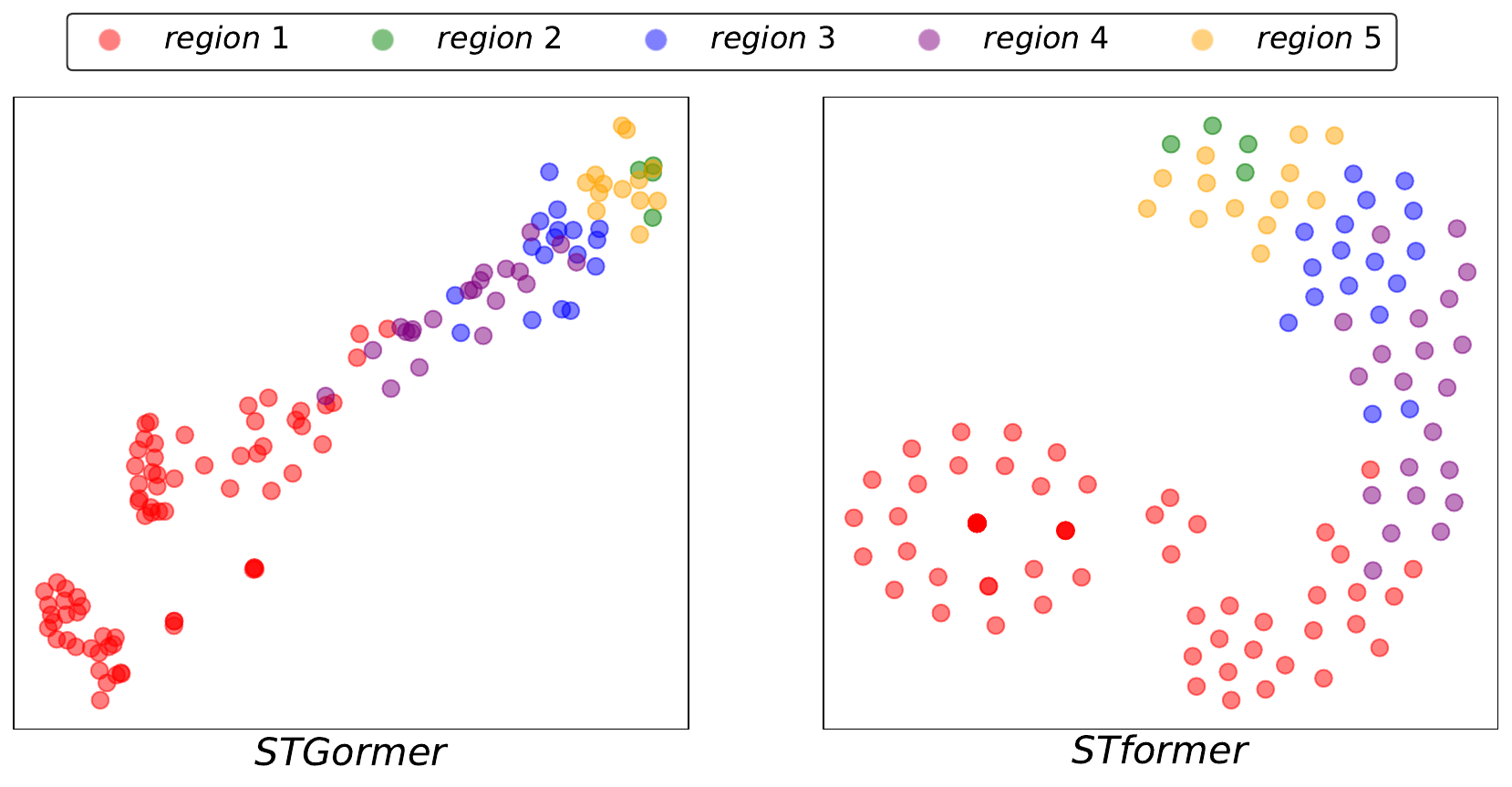}
        \caption{t-SNE visualization of embeddings.}
        \label{fig:5-case-tsne}
    \end{subfigure}
    \caption{Qualitative Study on NYCBike1.}
    \label{fig:5-case}
\end{figure}

\textbf{\textit{Visualization of Spatial Structural Embeddings.}} 
To demonstrate the capability of spatial structure embeddings in capturing essential spatial information, we employed the NYCBike1 dataset for our empirical studies, presented in a grid-based visualization. 
As illustrated in Figure~\ref{fig:5-case-s_in}, the embedding $s_{in}$ quantifies the significance of each node using degree centrality, enhancing the model's sensitivity to nodes of greater importance. 
Additionally, Figure~\ref{fig:5-case-sa_bias} visualizes the attention scores from a focal node (highlighted in red circle) to others within the spatial attention matrix $SA_{bias}$. This matrix, computed based on the shortest path distances between nodes, enables the transformer to incorporate graph structural information effectively. 
These embeddings are fundamental for modeling spatial relationships and are crucial for enhancing the performance of traffic forecasting task.

\textbf{\textit{Learned Spatio-temporal Representations.}} 
To further understand how the STGormer model achieves superior traffic prediction accuracy, we visualized its learned embeddings using t-SNE~\cite{van2008visualizing} for the NYCBike1 dataset. 
The embeddings represent the final outputs across all regions, with a clustering based on traffic data statistics (mean, median, standard deviation) to highlight spatial patterns among regions with heterogeneous data distributions. 
For comparison, we also analyzed embeddings from a simpler model, STformer, which solely utilizes spatial and temporal attention mechanisms. 
The t-SNE visualization, as depicted in Figure~\ref{fig:5-case-tsne}, shows that STGormer generates more compact clusters for regions within the same class and distinct separation between different classes. This distinction underscores STGormer's ability to recognize spatial heterogeneity and effectively transfer information across similar regions, thus facilitating more accurate predictions.

\section{Conclusion and Future Work}
In this paper, we introduce the Spatio-Temporal Graph Transformer (STGormer) to effectively capture spatio-temporal correlations and heterogeneity in traffic forecasting tasks. Our framework integrates temporal and spatial static context into the analysis of traffic flow data. The resultant fused representation is then fed into spatio-temporal transformers enhanced with Mixture of Experts (MoE), which incorporates spatial context in the attention layer and a spatio-temporal gating network. Finally, leveraging the learned spatio-temporal representation, our framework yields superior prediction results.
Extensive experiments conducted on real-world traffic datasets validate the efficacy of STGormer, outperforming existing traffic forecasting models.
In future research, we plan to further enhance and expand the application of STGormer to various spatio-temporal forecasting tasks, extending beyond traffic prediction. Additionally, we aim to explore other spatial structural embedding techniques to improve the transformer's ability to model spatial relationships, thereby boosting its overall forecasting performance.


\bibliographystyle{ACM-Reference-Format}
\bibliography{main}




\end{document}